\documentclass[10pt,conference,a4paper]{IEEEtran}

\usepackage{amsmath,graphicx}
\usepackage{mathtools}

\ifCLASSOPTIONcompsoc
 \usepackage[caption=false,font=normalsize,labelfont=sf,textfont=sf]{subfig}
\else
 \usepackage[caption=false,font=footnotesize]{subfig}
\fi

\usepackage{multirow}

\usepackage{rotating}

\DeclarePairedDelimiter\floor{\lfloor}{\rfloor}

\makeatletter
\newcommand{\linebreakand}{%
  \end{@IEEEauthorhalign}
  \hfill\mbox{}\par
  \mbox{}\hfill\begin{@IEEEauthorhalign}
}
\makeatother

\usepackage[top=0.705in, left=0.514in, bottom=1.705in, right=0.514in]{geometry}

\begin{document}

\title{Fast Implementation of 4-bit Convolutional Neural Networks for Mobile Devices}

\author{\IEEEauthorblockN{Anton Trusov}
\IEEEauthorblockA{Moscow Institute of Physics and Technology\\
Dolgoprudny, Russia\\
Smart Engines Service LLC\\
Moscow, Russia\\
Email: trusov.av@smartengines.ru
}
\and
\IEEEauthorblockN{Elena Limonova}
\IEEEauthorblockA{FRC CSC RAS\\
Moscow, Russia\\
Smart Engines Service LLC\\
Moscow, Russia \\
Email: limonova@smartengines.com}
\and
\IEEEauthorblockN{Dmitry Slugin}
\IEEEauthorblockA{FRC CSC RAS\\
Moscow, Russia\\
Smart Engines Service LLC\\
Moscow, Russia\\
Email: slugin@smartengines.com}
\linebreakand
\IEEEauthorblockN{Dmitry Nikolaev}
\IEEEauthorblockA{Institute for Information Transmission Problems RAS\\
Moscow, Russia\\
Smart Engines Service LLC\\
Moscow, Russia\\
Email: dimonstr@iitp.ru}
\and
\IEEEauthorblockN{Vladimir V. Arlazarov}
\IEEEauthorblockA{FRC CSC RAS\\
Moscow, Russia\\
Smart Engines Service LLC\\
Moscow, Russia\\
Email: vva@smartengines.com}}
\maketitle
\IEEEpeerreviewmaketitle

\begin{abstract}
Quantized low-precision neural networks are very popular because they require less computational resources for inference and can provide high performance, which is vital for real-time and embedded recognition systems. However, their advantages are apparent for FPGA and ASIC devices, while general-purpose processor architectures are not always able to perform low-bit integer computations efficiently. 
The most frequently used low-precision neural network model for mobile central processors is an 8-bit quantized network. However, in a number of cases, it is possible to use fewer bits for weights and activations, and the only problem is the difficulty of efficient implementation.
We introduce an efficient implementation of 4-bit matrix multiplication for quantized neural networks and perform time measurements on a mobile ARM processor. It shows 2.9 times speedup compared to standard floating-point multiplication and is 1.5 times faster than 8-bit quantized one. We also demonstrate a 4-bit quantized neural network for OCR recognition on the MIDV-500 dataset. 4-bit quantization gives 95.0\% accuracy and 48\% overall inference speedup, while an 8-bit quantized network gives 95.4\% accuracy and 39\% speedup. The results show that 4-bit quantization perfectly suits mobile devices, yielding good enough accuracy and low inference time.

\end{abstract}
\section{Introduction}
\label{sec:intro}
Nowadays, convolutional neural networks (CNNs) are widely used to solve pattern recognition, segmentation, object detection, and other problems \cite{bezmaternykh2019unetbin, laroca2018robust, lienhart2002localizing, robles2018convolutional}. Billions of mobile and IoT devices generate an enormous amount of new data for those tasks. So the traditional paradigm of cloud computing, when this data is stored at megascale datacenters and processed in the network core, causes significant network load and thus become obsolete. New paradigms tend to push computing tasks and services from the network core to the network edge (edge computing and edge intelligence)~\cite{zhou2019edge} and even to end devices. Unfortunately, small computers (IoT or mobile devices like smartphones) on the edge of the network have limited computational and memory resources compared to powerful servers. It strongly restricts the use of deep convolutional neural networks, which generally require a vast amount of memory and perform billions of multiplications. Moreover, real-time recognition systems (for example, \cite{usilin2020fast, bulatov2019on, povolotskiy2019dynamic}) impose even more time constraints on CNNs.

There are several ways for CNNs to satisfy the efficiency constraints of mobile devices. One way is to simplify neural network: use pruning \cite{molchanov2016pruning} to remove non-informative weights or knowledge distillation \cite{hinton2015distilling} to transfer knowledge to smaller network, train neural network with early exit \cite{teplyakov2020complexityaware}. Those methods modify or completely change the neural network, but we also can accelerate inference while preserving network architecture.

As was mentioned in \cite{dukhan2019indirect} direct convolution algorithm is extremely ineffective, compared to GEMM-based algorithms, which convert the convolution problem into a GEMM (Matrix-Matrix Multiplication) problem. Therefore a significant part of computations during inference of CNNs is matrix multiplication. Of course, to perform fast matrix multiplication on specific devices, one can use optimized libraries and BLAS packages. They provide efficient linear algebra algorithms and are available for many computing architectures. They suit perfectly for a fast inference of standard floating-point neural networks (Eigen library  \cite{guennebaud2010eigen}, for example). The next speed up step is to approximate floating-point matrices with low-bit integer ones. This process is called quantization, and neural networks that benefit from such approximations are called quantized neural networks (QNNs). The critical difference between the multiplication of low-bit matrices and floating-point ones is that low-bit multiplication should be widening. Therefore, we can not use standard methods and need special libraries and frameworks. For example, Google's gemmlowp library \cite{jacob2017gemmlowp} provides low-precision matrix multiplication for 8-bit QNNs. They use 8-bit unsigned integers for inputs and accumulate the result into 32-bit integers. In Facebook's QNNPACK \cite{dukhan2018qnnpack}, authors also use 8-bit unsigned integers for inputs but re-quantize 32-bit intermediates to get 8-bit outputs.

Of course, there are many QNNs, which use 8-bit~\cite{gysel2016hardware}, 4-bit \cite{zhuang2018towards}, 2-bit ternary \cite{alemdar2017ternary} even binary \cite{zhuang2019structured} and other low-bit quantization schemes \cite{Yang_2019_CVPR}. 
The main problem of such approaches is that they can achieve high performance basically on special chips \cite{yu2016binary} or FPGA~\cite{liang2018fp} because modern central (CPU) and graphical (GPU) processors do not provide enough commands to work with separate bits: they only allow to access 8-bit (or multiple of 8-bit) blocks and perform computations on them. Therefore, efficient QNN implementations for CPUs mainly do not use less than 8-bit matrices and highly specialize for a target CPU architecture to achieve inference time speed up \cite{limonova2020special}.



In this paper, we provide an algorithm for fast inference of 4-bit quantized neural network on CPU. As was proven before, CNN models with 4-bit quantization are viable, and even that with just a few-percent accuracy degradation, retraining CNN models may be unnecessary for 4-bit quantization \cite{banner2019post}. But to our best knowledge, no research was focused on an efficient implementation of 4-bit QNNs. 
This implementation is needed to exploit the full potential of CNNs in the paradigm of edge intelligence and the presence of performance and memory limitations.

This paper is organized as follows.
\begin{enumerate}
  \item In section \ref{sec:mul}, we describe frequently used linear quantization scheme and matrix multiplication for this scheme. Then we provide a novel effective algorithm of matrix multiplication for 4-bit quantized values, especially designed and optimized for ARM CPU architecture, which is commonly used in mobile devices. 
  \item In section \ref{sec:qnn} at first, we describe the 4-bit quantized convolution layer based on our multiplication algorithm and point out its pros and cons. Then we explain how to include those layers in CNN (which does not have to be necessarily fully quantized). Finally, we note a possible way to train QNN with our 4-bit quantized layers.
  \item In section \ref{sec:exp}, we experimentally measure the efficiency of our algorithm. Initially, we compare the run-time of our 4-bit quantized matrix multiplication to an 8-bit multiplication similar to one from the gemmlowp library, traditional floating-point, and 32-bit integer matrix multiplication. Then we show the computational efficiency of 4- and 8-bit QNNs with quantized convolutional layers compared to traditional floating-point CNNs and naive implementation of QNN based on 32-bit integer matrix multiplication. In the last experiment, we performed character recognition of the Machine-Readable Zone of documents from the MIDV-500 dataset \cite{Arlazarov2018MIDV500AD}.
  \item In section \ref{sec:disc}, we discuss possible applications of our 4-bit QNN and future work on other low-bit QNN implementations. 
  \item And in section \ref{sec:conc}, we summarize results achieved in this paper.
  \end{enumerate}

\section{Quantized matrix multiplication}
\label{sec:mul}
\subsection{Quantization scheme}
\label{ssec:mul:quant}

In the paper we use the linear quantization method:

\begin{equation}
    \label{eq:quant}
    \begin{split}
        \hat{w_i} &=  \floor*{\frac{w_i}{s}} - z \\
        s &= \dfrac{\max(\max_i{w_i}, 0) - \min(\min_i{w_i}, 0)}{2^p - 1} \\
        z &= \min(\min_i{w_i}, 0),
    \end{split}
\end{equation}
 where $\hat{w_i}$ denotes quantized values, $w_i$ are floating-point values, $s$ is scale factor, $z$ is a zero-point (offset), $p$ is a number of bits used in quantized values (in our case it is 4, in case of gemmlowp library -- 8). 
 
\subsection{Quantized multiplication}
\label{ssec:mul:mul}
 
Let us consider the quantized approximation of matrix multiplication $R = W X$:
 
 \begin{equation}
    \label{eq:quant_mul}
     \begin{split}
        &r_{ij} = \sum_{k = 1}^D w_{ik} x_{kj} \approx \sum_{k = 1}^D s_w(\hat{w}_{ik} - z_w) s_x(\hat{x}_{kj} - z_x) \\
        &= s_w s_x \Big(\sum_{k = 1}^D \hat{w}_{ik} \hat{x}_{kj} - z_w \sum_{k = 1}^D \hat{x}_{kj} -  z_x \sum_{k = 1}^D \hat{w}_{ik} + D z_x z_w   \Big) 
     \end{split}
 \end{equation}  
 where $r_{ij}$ denotes values of $R$ matrix, $w_{ik}$ and $x_{kj}$ are values of $W$ and $X$ matrices, $\hat{w}_{ik}$ and $\hat{w}_{ik}$ are their quantized approximations due to (\ref{eq:quant}), $s_w$ and $s_x$ are scale factors, $z_w$ and $z_x$ are zero-points and $D$ is a depth of multiplication (number of columns in left matrix ($W$) and rows in right matrix($X$)). The first term in (\ref{eq:quant_mul}) is a result of matrix multiplication of quantized matrices, the second is a sum over rows of quantized right matrix multiplied by zero-point of left matrix and can be effectively computed while unrolling activation matrix, the third term is the sum over columns of the left matrix multiplied by zero-point of the right matrix, and the last term is simply a constant. In QNN $W$ is a matrix of weights, so its quantized approximation, scale factor, zero point and sums over columns can be computed only once to speed up inference. 
 
 For 8-bit quantization $\hat{w}_{ik}$ and $\hat{x}_{kj}$ are unsigned 8-bit integers and thus have values in range $[0, \dots, 255]$, in process of multiplication they are zero-expanded to 16-bit integers and their sums (first term of (\ref{eq:quant_mul})) are stored in 32-bit integer accumulators. For 4-bit quantization (in range $[0 \dots, 15]$) zero-expanded to 8-bit with 16-bit accumulators.

\subsection{Quantized multiplication kernel}
\label{ssec:mul:ker}
\begin{figure}[ht]
        \centering
        \subfloat[Smaller kernel] {\includegraphics[height=4.5cm]{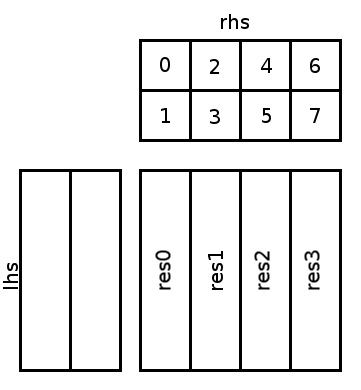}%
        \label{fig:ker:small}}
        \vfil
        \subfloat[Bigger kernel] {\includegraphics[height=9.cm]{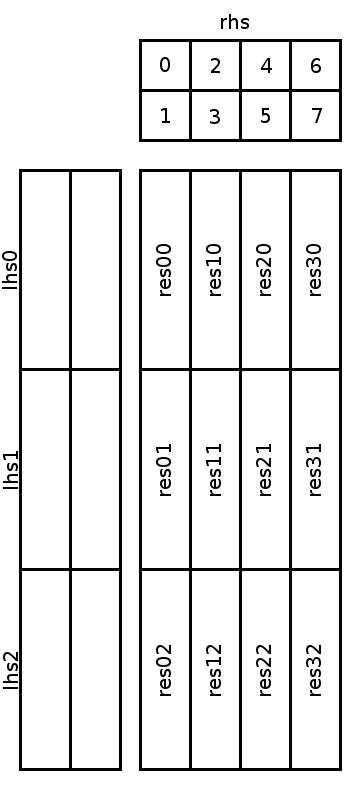}%
        \label{fig:ker:big}}
        \caption{Kernel layout. \textit{rhs} is a block of a right matrix and consists of 8 8-bit values. \textit{lhs} is a 128-bit block of a left matrix, storing 16 8-bit values. Each lower and upper parts of registers are separated with vertical line. \textit{res} are 128-bit blocks of a result matrix, storing 8 16-bit values each.}
        \label{fig:ker}
    \end{figure}

As mentioned above, to speed up neural network computations we need effective matrix multiplication. To achieve it we take advantage of SIMD-extension of ARM processor, which allows us to compute the same instruction on several values at once. ARM SIMD stores values in 128 or 64-bit registers. According to ARM architecture reference manual \cite{ARM2014} on ARM-v7 there are 16 128-bit registers which can be viewed as 32 64-bit registers. Knowing that we construct a multiplication kernel (a function that simultaneously computes several values of result matrix). It takes blocks from left and right matrices and computes a block of the result values which are then added to the current value of accumulators. In Fig.\ref{fig:ker} kernel layout is shown. We use 2 kernels: bigger one for a major part of a matrix and smaller for a "tail" of a matrix that can not be processed by a big kernel. The rest of the matrix is processed without SIMD acceleration. We use 16-bit accumulators, so our kernel sizes are $24 \times 4$ and $8 \times 4$ for bigger and smaller kernels respectively.  

\begin{figure}[ht]
        \centering
        \subfloat[RHS]{\includegraphics{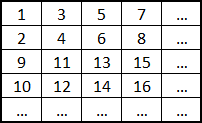}%
        \label{fig:pack:rhs}}
        \vfil
        \subfloat[LHS]{\includegraphics{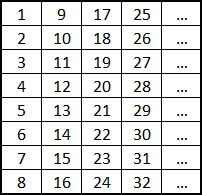}%
        \label{fig:pack:lhs1}}
        \caption{Order or values of right (\ref{fig:pack:rhs}) and left (\ref{fig:pack:lhs1}) matrices in temporal buffers}
        \label{fig:pack}
    \end{figure}

For kernel to work efficiently it needs to minimize the number of loads from memory, so before multiplication itself values from the left and right matrices packed in temporal buffers in a specific order. In our implementation values of the right matrix are stored in temporal buffer \texttt{RHS} so the first 2 values from 1$^\text{st}$, 2$^\text{nd}$, 3$^\text{d}$ and 4$^\text{th}$ column are followed by the second 2 values from the same columns, than 3$^\text{d}$, etc. If the number of columns is not divided by 4, or the number of rows is not divided by 2 the rest of the matrix is not packed into a temporal buffer and is processed without SIMD acceleration. The order of values in the temporal buffer of left matrix \texttt{LHS} depends on kernel size: for smaller kernel first 8 values from the 1$^\text{st}$ column are followed by the first 8 values from 2$^\text{nd}$, 3$^\text{d}$, etc. For bigger kernel its values 1-8 from 1$^\text{st}$ column 1-8 from 2$^\text{nd}$, 9-16 from 1$^\text{st}$, 9-16 from 2$^\text{nd}$, 17-24 from 1$^\text{st}$ 17-24 from 2$^\text{nd}$ than go 3$^\text{d}$ and 4$^\text{th}$ columns, etc. Again if the number of rows is not divided by the number of rows in kernel or number of columns is not divided by 2 is not packed into the temporal buffer and is processed without SIMD acceleration. Fig. \ref{fig:pack} illustrates how values from right and left matrices are stored in \texttt{RHS} and  \texttt{LHS} buffers, in case of smaller kernel.

 One step of matrix multiplication is shown in Fig.\ref{fig:ker} registers are denoted as \texttt{res} for 16-bit unsigned integer accumulators, \texttt{rhs} and \texttt{lhs} for blocks of 8 4-bit integers from right and left matrices, zero-extended to 8 bits. Registers \texttt{rhs} and \texttt{lhs} sequentially load values from buffers \texttt{RHS} and \texttt{LHS}, nubers denote order of elements in original matrix. This way we need only one memory call to load 64-bit \texttt{rhs} register and another one (or three for bigger kernel) to load 128-bit \texttt{lhs} registers. When all registers are loaded, each value of \texttt{rhs} register is duplicated and stored in another 64-bit register using VDUP\_LANE. The result is multiplied by corresponding half of \texttt{lhs} register and added to \texttt{res} register with a single VMAL instruction. After the summation over depth dimension is finished \texttt{res} registers are stored back into a memory.

For smaller kernel pseudocode of our method is given below:

\texttt{pack right matrix into RHS}

\texttt{pack left matrix into LHS}

\texttt{for j in {0, \dots, cols / nr}}

\hspace{0.3 cm}\texttt{\{res0 ... res3\} $\leftarrow$ next result block}

\hspace{0.3 cm}\texttt{for i in {0, \dots, rows / mr}}

\hspace{0.6 cm}\texttt{for k in {0, \dots, depth / 2}}

\hspace{0.9 cm}\texttt{lhs $\leftarrow$ next block from LHS}

\hspace{0.9 cm}\texttt{rhs $\leftarrow$ next block from RHS}

\hspace{0.9 cm}\texttt{VMAL(dst0, LOW(lhs), rhs[0]);}

\hspace{0.9 cm}\texttt{VMAL(dst0, HIGH(lhs), rhs[1]);}

\hspace{2.1 cm}\texttt{...}

\hspace{0.9 cm}\texttt{VMAL(dst3, LOW(lhs), rhs[6]);}

\hspace{0.9 cm}\texttt{VMAL(dst3, HIGH(lhs), rhs[7]);}

\hspace{0.6 cm}\texttt{result block $\leftarrow$ \{res0 ... res3\}}

Here notation \texttt{LOW} and \texttt{HIGH} stands for getting lower or higher half of a register (which does not generate any instructions on assembler level) and \texttt{rhs[n]} getting and duplicating \texttt{n}'th bit of \texttt{rhs} using VDUP\_LANE instruction, \texttt{cols}, \texttt{rows} and \texttt{depth} represent dimensions on matrices, \texttt{nr}, \texttt{mr} are height and width of a kernel (in our case they are 8 and 4 and 24 and 4 for smaller and bigger kernels respectively). Algorithm for a bigger kernel is similar to this one but uses more registers.

The bigger kernel uses 15 of 16 128-bit ARM vector registers and one 64-bit, in this way we minimize calls to memory and a total number of instructions: in inner loop we upload only 3-128 bit registers and one 64 bit and use 24 VMAL and VDUP\_LANE instructions to compute 384 multiplications and additions.

It is worth mentioning that 8-bit quantized and traditional floating-point matrix multiplication algorithms store result in 32-bit values (4 per 128-bit SIMD registers), and thus require twice as many multiplications of SIMD registers to compute result matrix.

\section{QNN model}
\label{sec:qnn}

  Based on proposed quantization scheme (\ref{eq:quant}) and algorithm of matrix multiplication, described above we propose QNN model.

  \subsection{Quantized convolution layer}
  \label{ssec:qnn:conv}
  Let us consider
  \begin{equation}
     \label{eq:quant_res}
      \hat{r}_{ij} = \sum_{k = 1}^D \hat{w}_{ik} \hat{x}_{kj} - z_w  \sum_{k = 1}^D  \hat{x}_{kj} -  z_x \sum_{k = 1}^D \hat{w}_{ik} +  D z_x z_w,
  \end{equation}
  where deffinition are the same as in (\ref{eq:quant_mul}). Than
  \begin{equation}
     \label{eq:quant_res_cast}
         r_{ij} = s_w s_x \hat{r}_{ij}.
  \end{equation}
 
  So we can treat $\{\hat{r}_{ij}\}$ as quantized matrix with the zero-point $z_r = 0$ and the scale factor $s_r = s_w s_x$. That is why we do not convert the quantized result back to 32-bit floating-point values but preserve it as is in 16-bit integer form for 4-bit quantization (and 32-bit integer for 8-bit quantization), then pass it through activation function to the next layer where it is quantized to 4-bit (8-bit) once again. If that quantization results in $s_x^*$ scale factor, than total scale factor will be $s = s_x^* s_r$.
 
  As we can see all terms in (\ref{eq:quant_mul}) are signed integers. In our case they are signed 16-bit integers, which means that first term should be less than $2^{15} - 1$ to avoid integer overflow. Due to~(\ref{eq:quant}) $\hat{w}_{ik}$ and $\hat{x}_{kj}$ are not greater than $15$, so their product is not greater than $225$. That gives us theoretical constraint on depth of matrix multiplication: $D \leq \floor*{\dfrac{2^{15} - 1}{225}} = 145$.
 
  In case of convolution $D = k_h k_w c$, where $k_h$ and $k_w$ are convolution kernel height and width, $c$ is a number of channels in input of current layer. So for $q$-bit quantization with $p$-bit accumulators we obtain a constraint on number of input channels:
 
  \begin{equation}
     \label{eq:quant_ch_constraint1}
         c \leq \floor*{\frac{2^{p-1} - 1}{(2^q - 1)^2 }} \frac{1}{ k_h k_w}.
  \end{equation}
 
  This gives us 145 channels for $1\times1$ convolution kernel, 16 channels for $3\times3$ kernel, and only 5 channels for $5\times5$ convolutions. To weaken this constraint one can compute the first term in unsigned 16-bit integers than zero expand it up to 32-bits and compute (\ref{eq:quant_res}) in 32-bit signed integers. This way first term should be less than $2^{16} - 1$, and so $ D~\leq~\floor*{\dfrac{2^{16} - 1}{225}} = 291.$
  
  However, the situation, when all quantized values in one row of unrolled filter matrix equal to $15$, is not likely to happen in practice, so one can try to use more channels.

 \subsection{Quantized network}
 \label{ssec:qnn:qnn}
 
 Although quantized integer computations are faster then floating-point, mobile CPUs do not constrain us to use only integer operations the way it is done in end-to-end QNNs \cite{Li_2019_CVPR}. To achieve better inference results, we can preserve some layers not quantized like in \cite{zhou2016dorefa}, where authors do not quantize the first and last layers of a network. In our experiment, we do not quantize only the last one. So our QNN model consists of quantized layers (Q) and not quantized layers (F) with floating-point weights and activations. 
 
 For each quantized layer, we convert weights matrix (unrolled for convolution layers) according to (\ref{eq:quant}) and save scale factor $s_w$, zero point $z_w$ and sums of rows of this matrix $\sum_{k = 1}^D \hat{w}_{ik}$.
 
 During network inference, the quantized layer converts its input according to (\ref{eq:quant}) and returns a 16-bit integer activation (\ref{eq:quant_res}) vector and a scale factor, that can be used to convert the data back to floating-point. 
 
 To be more specific let us consider inputs (with conversions if needed) and outputs of F and Q layers, depending on the previous layer (input of a network is treated as F).
 
 \begin{itemize}
    \item \textbf{F $\rightarrow$ F}. \textbf{Input}: floating-point activation. \textbf{Output}: floating-point activation.
    
    \item \textbf{F $\rightarrow$ Q}. \textbf{Input}: floating-point activation. It is quantized to 4-bit integer with scale factor $s_x$. \textbf{Output}: 16-bit integer activation, scale factor $s = s_x s_w$. 
    
    \item \textbf{Q $\rightarrow$ Q}. \textbf{Input}: 16-bit integer activation, scale factor $s^*$. Activation is quantized to 4-bit integer with scale factor $s_x$. \textbf{Output}: 16-bit integer activation, scale factor $s = s_x s_w s^*$.
    
    \item \textbf{Q $\rightarrow$ F}. \textbf{Input}: 16-bit integer activation, scale factor $s^*$. Activation is converted back to floating-point by multiplication by $s^*$. \textbf{Output}: floating-point activation.
 
 \end{itemize}
 
 \subsection{Training and fine-tuning}
 \label{ssec:qnn:train}
 
 To train our 4-bit QNN we first train CNN with only floating-point weights. We can not simply quantize all its layers and apply such a network, because quantization error for 4-bit approximation is significant and rises dramatically with increasing of the layer number. To deal with this problem we apply layer-by-layer fine-tuning as described in \cite{ilin2017fast}: quantize the first layer, freeze its weights and retrain the rest of the network, then quantize and freeze the second layer and so one. The fully-connected layer remains not quantized.

\section{Experiments}
\label{sec:exp}

 \subsection{Matrix multiplication}
 \label{ssec:exp:mul}
 
 First we compare the time required to compute matrix multiplication of 32-bit floating-point, 32-bit integers, 8-bit integers and 4-bit unsigned integers values. For 4-bit integers we use the algorithm described in section \ref{sec:mul}, for 8-bit we use an algorithm similar to one from gemmlowp, and for floating-point and 32-bit integers an algorithm from Eigen library.  We randomly initialize matrices, compute matrix multiplication several times, take average time and repeat the experiment with different initialization until a result is stable enough. We take different sizes of matrices that are close to those which appear in compact lightweight CNNs.
 
 All tests are performed in a single thread of ODROID-XU4 single-board computer \cite{ODROID2017} with Samsung Exynos5422 ARM processor. 
 
 Results are provided in a Table \ref{tab:mat_mul}. We can see that multiplication of 32-bit integers and floating point values takes almost the same time, multiplication of 4-bit unsigned integers works approximately 2.4 times faster than that of 32-bit floating-point and 1.4 times faster compared to 8-bit unsigned integers for the smaller kernel (upper half of the Table). For the bigger kernel it is 2.9 and 1.5 times respectively (lower half of the Table). 
 
 \begin{table}[ht]
    \caption{Matrix multiplication timing}
    \label{tab:mat_mul}
    \begin{center}
    \begin{tabular}{|c|c|c|c|c|c|c|c|}
    \hline
    \begin{turn}{-90} \textbf{Height} \end{turn} &
    \begin{turn}{-90} \textbf{Width}\end{turn} &
    \begin{turn}{-90} \textbf{Depth} \end{turn} &
    \begin{turn}{-90} \textbf{Floating point 32-bit time}\end{turn} &
    \begin{turn}{-90} \textbf{Int 32-bit time}\end{turn} &
    \begin{turn}{-90} \textbf{Unsigned int 8-bit time}\end{turn} &
    \begin{turn}{-90} \textbf{Unsigned int 4-bit time}\end{turn} &
    \begin{turn}{-90} \textbf{Unit}\end{turn} \\
    
    \hline
    \multirow{9}{*}{8} 
        &\multirow{3}{*}{100} 
        &       10  
            & $8.3 $ 
            & $8.3 $ 
            & $4.6 $ 
            & $3.9 $ 
            & $\mu s$\\
            \cline{3-7}
        & &     40    
            & $21 $   
            & $21 $ 
            & $12 $ 
            & $8.4 $ 
            & $\mu s$\\
            \cline{3-7}
        & &     100   
            & $53 $ 
            & $52 $ 
            & $28 $ 
            & $19 $ 
            & $\mu s$\\
        \cline{2-7}
        &\multirow{3}{*}{400} 
        &       10  
            & $32 $ 
            & $33 $ 
            & $17 $ 
            & $14 $ 
            & $\mu s$\\
            \cline{3-7}
        & &     40    
            & $90 $  
            & $90 $ 
            & $50 $ 
            & $33 $ 
            & $\mu s$\\
            \cline{3-7}
        & &     100   
            & $0.21 $ 
            & $0.21 $ 
            & $0.15 $ 
            & $0.11 $ 
            & $ms$\\
        \cline{2-7}
        &\multirow{3}{*}{1600} 
        &       10  
            & $0.13 $ 
            & $0.14 $ 
            & $0.070 $ 
            & $0.058 $ 
            & $ms$\\
            \cline{3-7}
        & &     40    
            & $0.35 $ 
            & $0.36 $ 
            & $0.25 $ 
            & $0.19 $ 
            & $ms$\\
            \cline{3-7}
        & &     100   
            & $2.2 $ 
            & $2.4 $ 
            & $0.78 $ 
            & $0.55 $ 
            & $ms$\\
    \hline
    \multirow{9}{*}{24} 
        &\multirow{3}{*}{100} 
        &       10  
            & $15 $ 
            & $15 $ 
            & $9.1 $ 
            & $6.2 $ 
            & $\mu s$\\
            \cline{3-7}
        & &     40    
            & $44 $ 
            & $43 $ 
            & $24 $ 
            & $15 $ 
            & $\mu s$\\
            \cline{3-7}
        & &     100   
            & $0.11 $ 
            & $0.10 $ 
            & $0.054 $ 
            & $0.032 $ 
            & $ms$\\
        \cline{2-7}
        &\multirow{3}{*}{400} 
        &       10  
            & $62 $ 
            & $61 $ 
            & $35 $ 
            & $24 $ 
            & $\mu s$\\
            \cline{3-7}
        & &     40    
            & $0.18 $ 
            & $0.17 $ 
            & $0.096 $ 
            & $0.058 $ 
            & $ms$\\
            \cline{3-7}
        & &     100   
            & $0.44 $ 
            & $0.42 $ 
            & $0.24 $ 
            & $0.15 $ 
            & $ms$\\
        \cline{2-7}
        &\multirow{3}{*}{1600} 
        &       10  
            & $0.24 $ 
            & $0.24 $ 
            & $0.14 $ 
            & $0.097 $ 
            & $ms$\\
            \cline{3-7}
        & &     40    
            & $0.72 $ 
            & $0.71 $ 
            & $0.43 $ 
            & $0.28 $ 
            & $ms$\\
            \cline{3-7}
        & &     100   
            & $3.2 $  
            & $3.1 $ 
            & $1.2 $ 
            & $0.76 $ 
            & $ms$\\
    \hline
    \end{tabular}
    \end{center}
 \end{table}

 \subsection{QNN}
 \label{ssec:exp:qnn}
 
  Acceleration of matrix multiplication promises that 4-bit QNN will work significantly faster than CNN of the same architecture.
 
\begin{figure}[ht]
        \centering
        \subfloat[MRZ from German driving licence in MIDV-500] {\includegraphics[width=8.cm]{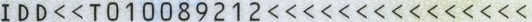}
        \label{fig:im:mrz}}
        \vfil
        \subfloat[Synthetic images] {
            \includegraphics[height=0.6cm]{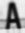}
            \hspace{1.cm}
            \includegraphics[height=0.6cm]{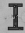}
            \hspace{1.cm}
            \includegraphics[height=0.6cm]{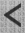}
            \hspace{1.cm}
            \includegraphics[height=0.6cm]{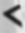}
            \label{fig:im:syn}
        }
        \vfil
        \subfloat[Real-world images] {
            \includegraphics[height=0.6cm]{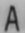}
            \hspace{1.cm}
            \includegraphics[height=0.6cm]{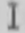}
            \hspace{1.cm}
            \includegraphics[height=0.6cm]{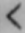}
            \hspace{1.cm}
            \includegraphics[height=0.6cm]{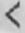}
            \label{fig:im:re}
        }
        \caption{Examples of images}
        \label{fig:im}
    \end{figure}

  We trained CNN that recognizes characters from the machine-readable zone (MRZ) of documents (for example Fig. \ref{fig:im:mrz}): its alphabet contains 36 characters (26 capital Latin letters, number 1-9 and character "$<$"). We treated the capital letter "O" and number "0" as the same character. Our network was trained on synthetic dataset, generated using method from \cite{chernyshova2020twostep} (Fig. \ref{fig:im:syn}). It was tested on images from MRZ of documents of MIDV-500 dataset \cite{Arlazarov2018MIDV500AD} (Fig. \ref{fig:im:re}). Because the dataset contains only text fields, not separate characters, we performed automatic segmentation using a segmentation network \cite{chernyshova2020twostep}.
  
  CNN architecture is provided in a Table \ref{tab:qnn_arch}, where Conv denotes a convolutional layer without bias or padding and FC a fully-connected layer. This network used fixed-size $25 \times 33$ grayscale images as an input. It contained 10892 trainable parameters. While converting to QNN all convolutional layers (8264 parameters) were quantized.
  
  \begin{table}[ht]
    \caption{CNN architecture}
    \label{tab:qnn_arch}
    \begin{center}
    \begin{tabular}{|l|l|l|l|}
        \hline
        \textbf{\#} & \textbf{Layer} & \textbf{Activation} & \textbf{Parameters} \\
            & \textbf{type} & \textbf{function} &  \\
        \hline
        1 & Conv & ReLU & 8 filters $5\times5$, stride $1\times1$ \\
        \hline
        2 & Conv & ReLU & 8 filters $3\times3$, stride $1\times1$ \\
        \hline
        3 & Conv & ReLU & 8 filters $3\times3$, stride $2\times2$ \\
        \hline
        4 & Conv & ReLU & 16 filters $3\times3$, stride $1\times1$ \\
        \hline
        5 & Conv & ReLU & 16 filters $3\times3$, stride $2\times2$ \\
        \hline
        6 & Conv & ReLU & 24 filters $3\times3$, stride $1\times1$ \\
        \hline
        7 & FC & SoftMax & 36 neurons \\
        \hline
    \end{tabular}
    \end{center}
 \end{table}

We also trained CNN and then converted it to QNN using 8-bit (QNN-8) and 4-bit (QNN-4) quantization. Than we compared their recognition accuracy on synthetic validation data from the training dataset and real-world data from MIDV-500 dataset, and also measured image recognition time and time per computation of all convolutions. Moreover we measured time of naive implementation of quantized network, that is based on multiplication of 32-bit integer matrices. This experiment was performed on the same ODROID-XU4 device. Results are presented in a Table \ref{tab:qnn_res}. To measure the time, we repeat the experiment several times until the deviation of the mean value descends to 1\%.  

\begin{table}[ht]
    \caption{Experiment results}
    \label{tab:qnn_res}
    \begin{center}
    \begin{tabular}{|c|c|c|c|c|}
        \hline
        \textbf{Model} & \textbf{Accuracy} & \textbf{Accuracy} & \textbf{Convolution} & \textbf{Total} \\
            & \textbf{synthetic}, \% & \textbf{MIDV}, \% & \textbf{time}, $ms$ & \textbf{time}, $ms$ \\
        \hline
        CNN   & 99.8 & 95.6 & 0.99 & 1.22 \\
        QNN-8 & 99.7 & 95.4 & 0.55 & 0.74 \\
        \textbf{QNN-4} & 99.2 & 95.0 & 0.45 & 0.63 \\
        QNN-32 & - & - & 1.16 & 1.47 \\
        \hline
    \end{tabular}
    \end{center}
 \end{table}

According to the results of experiment naive implementation of quantized neural network is even less efficient than floating point CNN, because additional time is required for quantization, while matrix multiplication gains no additional speedup. We can also see that both QNN-8 and proposed QNN-4 maintain a good quality of recognition and speed up computation by 39\% and 48\% respectively (QNN-4 is 1.17 times faster tham QNN-8). This is lower then speedup in the computation of convolutions (44\% and 54\%) because additional time is required for quantization of input of each layer.

\section{Discussion}
\label{sec:disc}

Our experiments prove that the efficient implementation of highly quantized neural networks is possible for CPUs. We have achieved almost 2 times speedup of CNN while preserving the number of trainable parameters with insignificant accuracy drop because of efficient usage of SIMD registers inside matrix multiplication and computing two times less register-wise multiplications. Though we experimented with 4-bit QNNs with classical linear quantization described in section \ref{ssec:mul:quant}, it will work with any QNN that relies on multiplication of 4-bit quantized values. For example, our algorithm could work with post-trained 4-bit QNNs \cite{banner2019post} that use analytical clipping for integer quantization (ACIQ) or with 4-bit QNNs obtained by outlier channel splitting (OCS) \cite{zhao2019improving}. However, our algorithm poses some constraints on the number of channels per layer in QNN (to avoid overflow of 16-bit accumulators), as was shown in section \ref{ssec:qnn:conv}. To overcome this limitation, one can use QNNs with depthwise separable convolutions \cite{chollet2017xception} or separated filters \cite{limonova2016computational} with our algorithm.

In this paper, we propose a novel algorithm for efficient inference of 4-bit QNNs on ARM processors, which allows us to use them on many mobile devices. That is a step that seems to be fruitful in the context of movement from the cloud-computing paradigm to edge and on-device intelligence. Though we only show the performance of 4-bit QNN in the relatively simple OCR task, it has already been proven that low-bit quantized CNNs demonstrate good scores in image classification tasks even on challenging datasets like ImageNet \cite{yang2019quantization, krizhevsky2012imagenet}.
So the further research steps in this direction can be:
\begin{enumerate}
\item providing algorithms for fast computation of QNNs with a different bit-width;
\item analyzing these algorithms in terms of computational efficiency, power usage, and memory load to choose the most suitable one for each case.
\end{enumerate}

\section{Conclusion}
\label{sec:conc}
This paper provides an implementation of an efficient matrix multiplication algorithm for 4-bit quantized matrices for mobile devices with ARM-based architecture. We experimentally prove its efficiency on Samsung Exynos5422 ARM processor: it works about 2.9 times faster than floating-point multiplication from Eigen library and 1.5 times faster than 8-bit quantized multiplication from gemmlowp library. 

Then we propose the 4-bit QNN model that benefits greatly from our algorithm of matrix multiplication. We demonstrate that this model is both viable and efficient. The real-world problem of OCR recognition on the MIDV-500 dataset demonstrates 95.0\% accuracy, while the floating-point network gives 95.6\% accuracy. Our network works 1.93 times faster than traditional CNN and 1.17 times faster than 8-bit QNN of the same architecture.

The proposed implementation allows creating applications for complex real-world tasks and satisfies modern smartphones and IoT devices' memory and computational constraints. Thus, this greatly expands the applicability of modern recognition methods and brings many quantization ideas closer to practical use in edge computing.

\section*{Acknowledgments}

This work is partially supported by Russian Foundation for Basic Research (projects 17-29-03240, 18-07-01384). 

\bibliographystyle{IEEEtran}
\bibliography{bibtex}

\begin{thebibliography}{10}
\providecommand{\url}[1]{#1}
\csname url@samestyle\endcsname
\providecommand{\newblock}{\relax}
\providecommand{\bibinfo}[2]{#2}
\providecommand{\BIBentrySTDinterwordspacing}{\spaceskip=0pt\relax}
\providecommand{\BIBentryALTinterwordstretchfactor}{4}
\providecommand{\BIBentryALTinterwordspacing}{\spaceskip=\fontdimen2\font plus
\BIBentryALTinterwordstretchfactor\fontdimen3\font minus
  \fontdimen4\font\relax}
\providecommand{\BIBforeignlanguage}[2]{{%
\expandafter\ifx\csname l@#1\endcsname\relax
\typeout{** WARNING: IEEEtran.bst: No hyphenation pattern has been}%
\typeout{** loaded for the language `#1'. Using the pattern for}%
\typeout{** the default language instead.}%
\else
\language=\csname l@#1\endcsname
\fi
#2}}
\providecommand{\BIBdecl}{\relax}
\BIBdecl

\bibitem{bezmaternykh2019unetbin}
P.~V. Bezmaternykh, D.~A. Ilin, and D.~P. Nikolaev, ``U-net-bin: hacking the
  document image binarization contest,'' \emph{Computer optics}, vol.~43,
  no.~5, pp. 825--832, 2019, dOI: 10.18287/2412-6179-2019-43- 5-825-832.

\bibitem{laroca2018robust}
R.~Laroca, E.~Severo, L.~A. Zanlorensi, L.~S. Oliveira, G.~R. Gon{\c{c}}alves,
  W.~R. Schwartz, and D.~Menotti, ``A robust real-time automatic license plate
  recognition based on the yolo detector,'' in \emph{2018 International Joint
  Conference on Neural Networks (IJCNN)}.\hskip 1em plus 0.5em minus
  0.4em\relax IEEE, 2018, pp. 1--10.

\bibitem{lienhart2002localizing}
R.~Lienhart and A.~Wernicke, ``Localizing and segmenting text in images and
  videos,'' \emph{IEEE Transactions on circuits and systems for video
  technology}, vol.~12, no.~4, pp. 256--268, 2002.

\bibitem{robles2018convolutional}
A.~Robles-Kelly and R.~Wei, ``A convolutional neural network for pixelwise
  illuminant recovery in colour and spectral images,'' in \emph{2018 24th
  International Conference on Pattern Recognition (ICPR)}.\hskip 1em plus 0.5em
  minus 0.4em\relax IEEE, 2018, pp. 109--114.

\bibitem{zhou2019edge}
Z.~Zhou, X.~Chen, E.~Li, L.~Zeng, K.~Luo, and J.~Zhang, ``Edge intelligence:
  Paving the last mile of artificial intelligence with edge computing,''
  \emph{Proceedings of the IEEE}, vol. 107, no.~8, pp. 1738--1762, 2019.

\bibitem{usilin2020fast}
S.~A. Usilin, P.~V. Bezmaternykh, and V.~V. Arlazarov, ``Fast approach for qr
  code localization on images using viola-jones method,'' in
  \emph{International Conference on Machine Vision (ICMV)}, vol. 11433, no.
  11433 3G.\hskip 1em plus 0.5em minus 0.4em\relax SPIE, Jan. 2020, pp. 1--9,
  dOI: 10.1117/12.2559386.

\bibitem{bulatov2019on}
K.~Bulatov, N.~Razumnyi, and V.~V. Arlazarov, ``On optimal stopping strategies
  for text recognition in a video stream as an application of a monotone
  sequential decision model,'' \emph{International Journal on Document Analysis
  and Recognition (AR)}, vol.~22, no.~3, pp. 303--314, 2019, dOI:
  10.1007/s10032-019-00333-0.

\bibitem{povolotskiy2019dynamic}
M.~A. Povolotskiy, D.~V. Tropin, T.~S. Chernov, and B.~I. Savelev, ``Dynamic
  programming approach to textual structured objects segmentation in images,''
  \emph{ITiVS}, vol.~69, no.~3, pp. 66--78, 2019, dOI: 10.14357/20718632190306.

\bibitem{molchanov2016pruning}
P.~Molchanov, S.~Tyree, T.~Karras, T.~Aila, and J.~Kautz, ``Pruning
  convolutional neural networks for resource efficient inference,'' in
  \emph{International Conference on Learning Representations (ICLR)}, 2016.

\bibitem{hinton2015distilling}
G.~Hinton, O.~Vinyals, and J.~Dean, ``Distilling the knowledge in a neural
  network,'' \emph{arXiv preprint arXiv:1503.02531}, 2015.

\bibitem{teplyakov2020complexityaware}
L.~Teplyakov, S.~Gladilin, and E.~Shvets, ``Complexity-aware loss function for
  fast neural networks with early exits,'' in \emph{International Conference on
  Machine Vision (ICMV)}, vol. 11433, no. 114333I.\hskip 1em plus 0.5em minus
  0.4em\relax SPIE, Jan. 2020, dOI: 10.1117/12.2557077.

\bibitem{dukhan2019indirect}
M.~Dukhan, ``The indirect convolution algorithm,'' \emph{arXiv preprint
  arXiv:1907.02129}, 2019.

\bibitem{guennebaud2010eigen}
\BIBentryALTinterwordspacing
G.~Guennebaud, B.~Jacob \emph{et~al.}, ``Eigen: a c++ linear algebra library,''
  2010. [Online]. Available: \url{http://eigen.tuxfamily.org}
\BIBentrySTDinterwordspacing

\bibitem{jacob2017gemmlowp}
\BIBentryALTinterwordspacing
B.~Jacob \emph{et~al.}, ``gemmlowp: a small self-contained low-precision gemm
  library,'' 2017. [Online]. Available:
  \url{https://github.com/google/gemmlowp}
\BIBentrySTDinterwordspacing

\bibitem{dukhan2018qnnpack}
\BIBentryALTinterwordspacing
M.~Dukhan, Y.~Wu, and H.~Lu, ``Qnnpack: open source library for optimized
  mobile deep learning,'' 2018. [Online]. Available:
  \url{https://code.fb.com/ml-applications/qnnpack/}
\BIBentrySTDinterwordspacing

\bibitem{gysel2016hardware}
P.~Gysel, M.~Motamedi, and S.~Ghiasi, ``Hardware-oriented approximation of
  convolutional neural networks,'' \emph{arXiv preprint arXiv:1604.03168},
  2016.

\bibitem{zhuang2018towards}
B.~Zhuang, C.~Shen, M.~Tan, L.~Liu, and I.~Reid, ``Towards effective
  low-bitwidth convolutional neural networks,'' in \emph{Proceedings of the
  IEEE Conference on Computer Vision and Pattern Recognition (CVPR)}, 2018, pp.
  7920--7928.

\bibitem{alemdar2017ternary}
H.~Alemdar, V.~Leroy, A.~Prost-Boucle, and F.~P{\'e}trot, ``Ternary neural
  networks for resource-efficient ai applications,'' in \emph{2017
  International Joint Conference on Neural Networks (IJCNN)}.\hskip 1em plus
  0.5em minus 0.4em\relax IEEE, 2017, pp. 2547--2554.

\bibitem{zhuang2019structured}
B.~Zhuang, C.~Shen, M.~Tan, L.~Liu, and I.~Reid, ``Structured binary neural
  networks for accurate image classification and semantic segmentation,'' in
  \emph{Proceedings of the IEEE Conference on Computer Vision and Pattern
  Recognition (CVPR)}, 2019, pp. 413--422.

\bibitem{Yang_2019_CVPR}
J.~Yang, X.~Shen, J.~Xing, X.~Tian, H.~Li, B.~Deng, J.~Huang, and X.-s. Hua,
  ``Quantization networks,'' in \emph{The IEEE Conference on Computer Vision
  and Pattern Recognition (CVPR)}, June 2019.

\bibitem{yu2016binary}
S.~Yu, Z.~Li, P.-Y. Chen, H.~Wu, B.~Gao, D.~Wang, W.~Wu, and H.~Qian, ``Binary
  neural network with 16 mb rram macro chip for classification and online
  training,'' in \emph{2016 IEEE International Electron Devices Meeting
  (IEDM)}.\hskip 1em plus 0.5em minus 0.4em\relax IEEE, 2016, pp. 16--2.

\bibitem{liang2018fp}
S.~Liang, S.~Yin, L.~Liu, W.~Luk, and S.~Wei, ``Fp-bnn: Binarized neural
  network on fpga,'' \emph{Neurocomputing}, vol. 275, pp. 1072--1086, 2018.

\bibitem{limonova2020special}
E.~E. Limonova, M.~I.-O. Neyman-Zade, and V.~L. Arlazarov, ``Special aspects of
  matrix operation implementations for low-precision neural network model on
  the elbrus platform,'' \emph{Bulletin of the South Ural State University.
  Ser. Mathematical Modelling, Programming \& Computer Software}, vol.~13,
  no.~1, pp. 118--128, 2020, dOI: 10.14529/mmp200109.

\bibitem{banner2019post}
R.~Banner, Y.~Nahshan, and D.~Soudry, ``Post training 4-bit quantization of
  convolutional networks for rapid-deployment,'' in \emph{Advances in Neural
  Information Processing Systems}, 2019, pp. 7950--7958.

\bibitem{Arlazarov2018MIDV500AD}
V.~V. Arlazarov, K.~Bulatov, T.~Chernov, and V.~L. Arlazarov, ``Midv-500: A
  dataset for identity document analysis and recognition on mobile devices in
  video stream,'' \emph{Computer optics}, vol.~43, no.~5, pp. 818--824, 2019,
  dOI: 18287/2412-6179-2019-43-5-818-824.

\bibitem{ARM2014}
\BIBentryALTinterwordspacing
\emph{ARM architecture reference manual ARMv7-A and ARMv7-R edition}, 2014.
  [Online]. Available:
  \url{https://static.docs.arm.com/ddi0406/c/DDI0406C\_C\_arm
  \_architecture\_reference\_manual.pdf}
\BIBentrySTDinterwordspacing

\bibitem{Li_2019_CVPR}
R.~Li, Y.~Wang, F.~Liang, H.~Qin, J.~Yan, and R.~Fan, ``Fully quantized network
  for object detection,'' in \emph{The IEEE Conference on Computer Vision and
  Pattern Recognition (CVPR)}, June 2019.

\bibitem{zhou2016dorefa}
S.~Zhou, Y.~Wu, Z.~Ni, X.~Zhou, H.~Wen, and Y.~Zou, ``Dorefa-net: Training low
  bitwidth convolutional neural networks with low bitwidth gradients,''
  \emph{arXiv preprint arXiv:1606.06160}, 2016.

\bibitem{ilin2017fast}
D.~A. Ilin, E.~E. Limonova, V.~V. Arlazarov, and D.~P. Nikolaev, ``Fast integer
  approximations in convolutional neural networks using layer-by-layer
  training,'' in \emph{International Conference on Machine Vision (ICMV)},
  A.~Verikas, Ed., vol. 10341, no. 103410Q.\hskip 1em plus 0.5em minus
  0.4em\relax Bellingham, Washington 98227-0010 USA: SPIE, July 2017, pp. 1--5,
  dOI: 10.1117/12.2268722.

\bibitem{ODROID2017}
\BIBentryALTinterwordspacing
\emph{User monual ODROID-XU4}, 2017. [Online]. Available:
  \url{https://magazine.odroid.com/wp-content/uploads/odroid-xu4-user-manual.pdf}
\BIBentrySTDinterwordspacing

\bibitem{chernyshova2020twostep}
Y.~S. Chernyshova, A.~V. Sheshkus, and V.~V. Arlazarov, ``Two-step cnn
  framework for text line recognition in camera-captured images,'' \emph{IEEE
  Access}, vol.~8, pp. 32\,587--32\,600, 2020, dOI:
  10.1109/ACCESS.2020.2974051.

\bibitem{zhao2019improving}
R.~Zhao, Y.~Hu, J.~Dotzel, C.~De~Sa, and Z.~Zhang, ``Improving neural network
  quantization without retraining using outlier channel splitting,''
  \emph{arXiv preprint arXiv:1901.09504}, 2019.

\bibitem{chollet2017xception}
F.~Chollet, ``Xception: Deep learning with depthwise separable convolutions,''
  in \emph{Proceedings of the IEEE conference on computer vision and pattern
  recognition}, 2017, pp. 1251--1258.

\bibitem{limonova2016computational}
E.~Limonova, A.~Sheshkus, and D.~Nikolaev, ``Computational optimization of
  convolutional neural networks using separated filters architecture,''
  \emph{International Journal of Applied Engineering Research}, vol.~11,
  no.~11, pp. 7491--7494, 2016.

\bibitem{yang2019quantization}
J.~Yang, X.~Shen, J.~Xing, X.~Tian, H.~Li, B.~Deng, J.~Huang, and X.-s. Hua,
  ``Quantization networks,'' in \emph{Proceedings of the IEEE Conference on
  Computer Vision and Pattern Recognition}, 2019, pp. 7308--7316.

\bibitem{krizhevsky2012imagenet}
A.~Krizhevsky, I.~Sutskever, and G.~E. Hinton, ``Imagenet classification with
  deep convolutional neural networks,'' in \emph{Advances in neural information
  processing systems}, 2012, pp. 1097--1105.

\end{thebibliography}
\end{document}